# SplitGNN: Splitting GNN for Node Classification with Heterogeneous Attention

Xiaolong Xu[*]
yiyin.xxl@antgroup.com
Ant Group
Hangzhou, Zhejiang, China

Lingjuan Lyu
lingjuanlvsmile@gmail.com
Ant Group
Hangzhou, Zhejiang, China

Yihong Dong
dongyihongbill@gmail.com
Ant Group
Hangzhou, Zhejiang, China

Yicheng Lu
sier.lyc@antgroup.com
Ant Group
Hangzhou, Zhejiang, China

Weiqiang Wang
weiqiang.wwq@antgroup.com
Ant Group
Hangzhou, Zhejiang, China

Hong Jin
jinhong.jh@antgroup.com
Ant Group
Hangzhou, Zhejiang, China

## ABSTRACT

With the frequent happening of privacy leakage and the enactment of privacy laws across different countries, data owners are reluctant to directly share their raw data and labels with any other party. In reality, a lot of these raw data are stored in the graph database, especially for finance. For collaboratively building graph neural networks (GNNs), federated learning (FL) may not be an ideal choice for the vertically partitioned setting where privacy and efficiency are the main concerns. Moreover, almost all the existing federated GNNs are mainly designed for homogeneous graphs, which simplify various types of relations as the same type, thus largely limiting their performance. We bridge this gap by proposing a split learning-based GNN (SplitGNN), where this model is divided into two sub-models: the local GNN model includes all the private data related computation to generate local node embeddings, whereas the global model calculates global embeddings by aggregating all the participants' local embeddings. Our SplitGNN allows the isolated heterogeneous neighborhood to be collaboratively utilized. To better capture representations, we propose a novel Heterogeneous Attention (HAT) algorithm and use both node-based and path-based attention mechanisms to learn various types of nodes and edges with multi-hop relation features. We demonstrate the effectiveness of our SplitGNN on node classification tasks for two standard public datasets. Experimental results validate that our proposed SplitGNN significantly outperforms the state-of-the-art (SOTA) methods.

[*]Corresponding author

## 1 INTRODUCTION

GNNs have attracted great interest in a variety of applications. GNNs can roughly be classified into two classes based on the graph structure: *heterogeneous graph* and *homogeneous graph*. Generally, the heterogeneous graph contains multiple node- and relation-types. On the contrary, in the homogeneous graph, nodes are objects of the same entity type and links are relationships from the same relation type. For the homogeneous graph, various effective methods have been investigated. For example, GraphSAGE [27] can derive the information of neighbors using an aggregation function. Graph convolutional network (GCN) [13] conducts the average pooling for each node's neighbors and then uses both linear projection and nonlinear activation operations. Graph attention networks (GAT) [26] adopts an effective attention mechanism and achieves a more powerful representation empirically. However, these methods cannot perform well in heterogeneous graphs, where abundant features are lying on edges (*e.g.* view frequency, watch duration, and public year, *et al.*) [5, 23, 25, 28, 33]. On the other hand, training accurate GNN models requires a wealth of high-quality graph-structured data, including rich node features and complete adjacent information. However, in practice, due to business competition and regulatory restrictions, such information could possibly be isolated by different participants, who are unwilling to share their information, plaguing many practical applications, such as fraud detection over banks and social network recommendation over platforms. Such data isolation problems present a serious challenge for the development of GNNs.

Federated learning (FL) and split learning (SL) are two promising distributed machine learning (ML) approaches that have gained attention due to their inherent privacy-aware capabilities that allow participants to collaboratively learn models without disclosing their raw data. Several recent works have attempted to build federated GNNs when data are horizontally partitioned [22, 31]. However, few works have studied the problem of GNN when data are vertically partitioned, which popularly exists in practice. In a vertically partitioned setting, both features and edges are distributed across different participants. For example, assume there are three participants (A, B, and C) and they have four same nodes. The node features are vertically split, i.e., A has $f1$, $f2$, and $f3$, B has $f4$ and $f5$, and C has $f6$ and $f7$. Meanwhile, A, B, and C may have their own edges. For instance, A has social relation between nodes while B and C have the payment relation between nodes. We also assume A is the party that has the node labels. The problem becomes how to collaboratively build federated GNN models by using the graph data of $A$, $B$, and $C$.

In this paper, we take the vertically partitioned setting for example, and present how to collaboratively build GNN models by leveraging the privacy and efficiency advantage of split learning [18]. Unlike previous privacy-preserving ML models that assume only nodes are held by different parties but samples have no relationship, our task is more challenging because GNN relies on the samples' relationships, which are also kept by different participants. More recently, Zhou *et al.* [32] have attempted federated GNN models under the vertically setting, however, their work was tested on smallscale homogeneous graph datasets, limiting their practicability in large-scale industrial applications. Moreover, the graph topology was still exploited locally, the model performance may be substantially reduced when the dataset is largely decentralized. To date, a mature solution for federated GNN models under the vertically partitioned setting is still missing. To fill in this gap and facilitate modeling heterogeneous graphs in the vertically partitioned setting, in this paper, we propose a novel framework, i.e., SplitGNN, for node classification across multiple heterogeneous graphs by splitting the whole GNN across different participants.

In summary, the main contributions of this paper include: 1) We present a novel approach, named SplitGNN, which combines SL with FL and eliminates their inherent drawbacks. To learn various types of nodes and edges in the heterogeneous graphs under the vertically partitioned setting, we design a novel base model named HAT, which uses both node-based and path-based attention mechanisms with multi-hop relations.2) We evaluate the different interactive layer strategies for the server to operate local node embeddings from participants, then test the SplitGNN performance on different data distributions and analyze the communication effectiveness of our proposed method compared with federated learning.

## 2 SPLITGNN

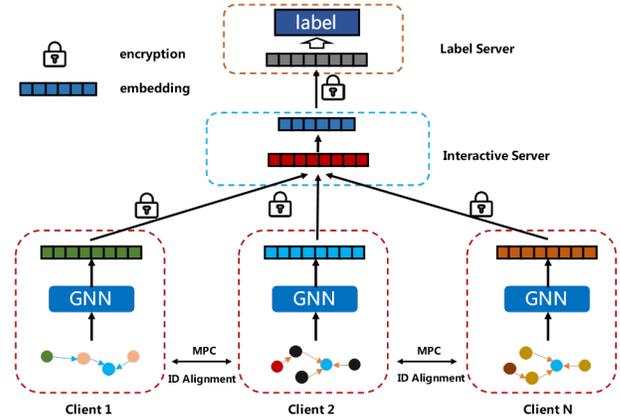

Figure 1: Workflow for SplitGNN.

### 2.1 Threat Model

The security model can be generally categorized into two types, i.e., the honest-but-curious (semi-honest) model and the malicious model. Although the semi-honest setting is less harsh than the malicious setting, it is more practical and has better efficiency than the latter. Hence, we consider semi-honest adversaries. That is, participants and the server strictly follow the protocol, but they also use all intermediate computation results to infer as much information as possible. We also assume that the server does not collude with any party. This security setting is commonly considered in many existing works [11]. We remark that this is a reasonable assumption since the server can be played by authorities such as governments or replaced by a trusted execution environment [6].

### 2.2 Overview of SplitGNN

In our framework, participants are willing to cooperatively train a global model with the aid of a server storing a fraction of the model, but are reluctant to directly share raw data and labels, because not only the raw samples (e.g., chest X-ray images) but also their groundtruth labels (e.g., lung cancer diagnosis) are privacy-sensitive. In actual operation, the label server is the same as one of the participants who want to enhance the classification model's capabilities with data from other participants. In an NN model, each raw sample is fed into the input layer, and its ground-truth label is compared with the model's prediction for the loss calculation at the output layer. Therefore, to preserve the privacy of each sample-and-label pair, both input and output layers should be stored by each participant, while the rest of the layers can be offloaded to the server, resulting in a tripartite SplitGNN. This is in stark contrast to the standard bipartite Split model where only the input layer is stored at each participant, while the remaining layers can be offloaded to the server.

In more detail, in our tripartite SplitGNN, the forward steps at each layer can be divided into three steps: it first calculates local embedding at each participant individually with private data. Then, the semi-honest server collects non-private local embeddings to compute global embedding. In the end, the server returns the final hidden layer to the party that has labels to compute prediction and loss. Participants and the server perform forward and back propagations to complete model training and prediction, during which the private data (i.e., features, edges, and labels) are always kept by the participants themselves. Compared with our SplitGNN, the other relevant work Zhou [32] exhibits the following weaknesses: 1) the full dataset is used in each iteration, which incurs much more communication cost; 2) data holders send their local node embeddings in plaintext form to the server, which may incur privacy leakage; 3) the server transmits the last hidden layer output in plaintext to the data holder who has the label, which also poses privacy issues.

## 2.3 ID alignment

The first step in collaborative modeling under vertical data split setting is secure entity alignment, also known as Private Set Intersection (PSI). That is, in each communication round, participants align their entities (nodes) without exposing those that do not overlap with each other. The server should record all the overlapped entities without knowing the details of any entity. Therefore, we need a privacy-preserving protocol to conduct ID alignment.

## 2.4 Local Node Embeddings

In SplitGNN, participants generate initial node embeddings using their own graph information, individually. With the generated node embeddings, one can do further tasks such as classification [12], link prediction [30], and recommendation [29] using deep neural networks.

For local node embeddings, similar to the existing GNNs, we perform multi-hop neighborhood aggregation on graphs using private edge information individually. Under the data isolated setting, neighborhood aggregation should be done by participants separately, rather than cooperatively, to protect the private edge information. This is because one may infer the neighborhood information of $v$ given the neighborhood aggregation results of $k$-hop ($h^k_v(i)$) and ($k+1$)-hop ($h^{k+1}_v(i)$), if neighborhood aggregation is done by partic-ipants together. A special case is $h^k_v(i) = h^{k+1}_v(i)$, where it is likely that $v$ is an isolated node in the graph of participant $i$.

For $\forall v \in V$ at each participant, the neighborhood aggregation is the same as the traditional GNN.

After participants generate local node embeddings, they need to send their local node embeddings to a semi-honest server for combination and further computations. Although the local node embeddings $h_v(i)$ hide raw information of local graphs, it may still encode the sensitive information of local graphs, hence participants need to either encrypt or perturb the extracted embeddings before sending to the server for further global computation.

## 2.5 Details of HAT

*2.5.1 Metapath Sampling With Relation Features.* In HAT, both pre-defined metapaths and original relations are taken to generate input subgraphs.

To capture more complex representations, when using pre-defined metapaths, we concatenate all the features of nodes and edges lying on the metapath. For the graph sampling process, all the original relations of neighbors will be considered to generate input subgraphs. By contrast, random sampling will miss some important information [1, 2, 14, 15], leading to weak model performance.

*2.5.2 Node Attention.*

$$h^l_i = W^{\tau,l} f^l_i + b_{\tau,l}, \quad r^{\rho,l}_{i,j} = W^{\rho,l} e^l_i + b_{\rho,l}, \quad h^{\rho,l}_i = \Phi(h^l_i, r^{\rho,l}_{i,j}) \quad (1)$$

where $f^l_i$ and $e^l_i$ refer to node features and edge features, $h^l_i$ and $r^{\rho,l}_{i,j}$ are latent vectors of $i$-th node and the edge which is the relation between the target node and $i$-th node in $l$-th layer of the $\rho$ relation, respectively. $W^{\tau,l} \in R^{D_n \times d}$ and $W^{\rho,l} \in R^{D_e \times d}$ stand for transformation weights of the node (type $\tau$) and edge (relation $\rho$) in the $l$-th layer respectively.

The function $\Phi$ represents the concatenation function which is used to combine node and edge latent vectors. In fact, other functions (*e.g.* linear transformation, element-wise addition) can be adopted as well. $h^{\rho,l}_i$ donates the node hidden status of $l$-th layer. The feature transformation stage will be processed in each subgraph $\rho$ for $N^{\rho}_i$. But for the target node, $h^{\rho,l}_i = h^l_i$.

$$z^{\rho,l}_i = F_{\rho}(h^{\rho,l}_i, h^{\rho,l}_j) \quad (2)$$

Where $F_{\rho}$ represents the aggregation function in the $\rho$ relation (or metapath). z donates node's hidden status embedding after node fusion. Here, we adopt the attention mechanism to integrate node embeddings into one vector.

$$z^{\rho,l}_i = \sigma\left( \Big\|_{m=1}^{M} \sum_{j \in N^{\rho}_i} \alpha^{\rho,l,m}_{i,j} h^{\rho,l}_j \right) \quad (3)$$

Here, $M$ is the multi-head number. In our work, we use Eq. (3) to aggregate node embeddings when processing the fusion stage. Where $\sigma$ is an active function. To consider the impact of different neighbor (middle neighbor) nodes on the target node, all neighbors have calculated attention coefficients in a given relation $\rho$.

$$\alpha^{\rho,l,m}_{i,j} = \frac{exp(h^{\rho,l}_i \cdot h^{\rho,l}_j)}{\sum_{k \in N^{\rho}_i} exp(h^{\rho,l}_i \cdot h^{\rho,l}_k)} \quad (4)$$

After this stage, every specific relation (metapath) would be simplified as one hidden state vector which contains all information of its subgraphs.

*2.5.3 Path Attention.* In the heterogeneous graph, there may exist multi-relations between two nodes. Based on this, path-level attention is needed which can reflect how the target node is impacted by different relations.

## 2.6 Server Aggregation

For the privacy-preserving aggregation of local node embeddings, since each local subgraph contains sensitive information about nodes, edges, attributes, and labels, the value of intermediate features has the potential risk to reveal sensitive information about the input data [17]. Henceforth, the intermediate representations should be transmitted in a secure and communication-efficient way. To provide the provable privacy guarantee, various privacy-preserving techniques can be considered, including secure aggregation [4], differential privacy (DP) [10]. To provide high security guarantees against the semi-honest server, we adopt homomorphic encryption (HE) [3, 7–9, 19, 20]. In this way, the server can only decrypt the aggregation of local embeddings. In terms of embedding aggregation, the combination strategy would be trainable in order to maintain high representational capacity. After receiving local node embeddings from all participants, the semi-honest server generates global node embeddings. Based on the aggregated representation, the server can conduct successive computations. We introduce three combination strategies, i.e., averaging, concatenation, and weighted averaging as follows. Note that adopting averaging or weighted averaging in contrast to direct concatenation allows the server's input dimension to remain fixed – independent of the number of participants.

Average. The average operator takes the elementwise average of the vectors in $(\{\mathbf{h}_v^K(i), \forall i \in \mathcal{P}\})$, assuming participants contribute equally to the global node embeddings, i.e., $h^K_v \leftarrow$ Average($\{h^K_v(1), h^K_v(2), \cdots, h^K_v(I)\}$) (6)

Concatenation. The concatenation operator can fully preserve local node embeddings learned from different participants. The operation can be written as below:

$$h^K_v \leftarrow \text{Concat}(\{h^K_v(1), h^K_v(2), \cdots, h^K_v(I)\}) \quad (7)$$

Weighted Average. The average strategy treats participants equally. In reality, the local node embeddings from different participants may contribute diversely to the global node embeddings. The weighted average operator aims to aggregate the embedding elements from participants through a regression model, whose parameters are learned intelligently during training. Let $\omega_i$ be the weight vector of local node embeddings from participant $i \in P$, then

$$h^K_v \leftarrow \omega_1 \odot \mathbf{h}_v^K(1) + \omega_2 \odot \mathbf{h}_v^K(2) ... + \omega_I \odot \mathbf{h}_v^K(I) \quad (8)$$

where $\odot$ is element-wise multiplication. Regression can handle the situation where the data quality and quantity of participants are different from each other.

## 3 EXPERIMENTS

### 3.1 Experimental Settings

We evaluate our SplitGNN on two real-world datasets (ACM and IMDB) with multiple classes.

Parameters Setting. We sample a batch of nodes ($\mathcal{B}$ = 512) in each communication round and report the performance after five communication rounds. For models on the server, we set the dropout rate to 0.3 and the network structure as ($d, d, |C|$), where $d \in \{32, 64, 128\}$ is the dimension of node embeddings and $|C|$ is the number of classes.

Baselines. 1) Standalone. Each participant trains its own GNN on their limited local data without any collaboration. 2) Entire. All participants pool their local data to a central server, which trains a global GNN on all the combined data.

### 3.2 Experimental Results

We first summarize the results in Table 1, where SplitGNN$_m$, SplitGNN$_c$, and SplitGNN$_w$ denote SplitGNN with Average, Concatenation, and Trainable Weighted Average combination strategies. It can be clearly observed that our SplitGNN significantly outperforms the GNNs by using the isolated data and has comparable performance with the traditional GNN by using the entire plain data insecurely. The reason is straightforward: the learned global node representation of SplitGNN is similar to that learned over the combined graph.

*3.2.1 Impact of the interactive layer strategy.* From Table 1, we find that the concatenation strategy performs the best generally. This is because the concatenation strategy can more fully integrate nodes' information coming from different clients.

*3.2.2 Impact of the number of participants.* We vary the number of participants in {2, 4, 8} and study the performance. In Table 2, we find that, as the number of participants increases, the accuracy of all models decreases, and the gaps widen with the increase of participants. We hypothesize this is because the neighborhood aggregation in SplitGNN is done by each participant individually for privacy concerns, and each participant will have fewer edge data when there are more participants since they split the original edge information evenly.

Table 1: Micro F1 Score of different base GNN models (Data is vertically partitioned (Ratio=5:5) among participants).

| Dataset | Strategy | GCN | GAT | HAT |
|---|---|---|---|---|
| ACM | Entire | 0.9007 | 0.9028 | 0.9158 |
| | Standalone_A | 0.7075 | 0.7134 | 0.7282 |
| | Standalone_B | 0.7087 | 0.7146 | 0.7307 |
| | SplitGNN_w | 0.8812 | 0.8837 | 0.8903 |
| | SplitGNN_c | 0.8827 | 0.8864 | 0.8984 |
| | SplitGNN_m | 0.8811 | 0.8816 | 0.8960 |
| IMDB | Entire | 0.5549 | 0.5396 | 0.5694 |
| | Standalone_A | 0.4273 | 0.4188 | 0.4429 |
| | Standalone_B | 0.4209 | 0.4185 | 0.4403 |
| | SplitGNN_w | 0.5385 | 0.5267 | 0.5567 |
| | SplitGNN_c | 0.5489 | 0.5251 | 0.5573 |

| | | | |
|---|---|---|---|
| SplitGNN_m | 0.5433 | 0.5235 | 0.5562 |

Table 2: Micro F1 Score of different base GNN models (Dataset: ACM, Strategy: concatenation, Ratio: equal size).

| Clients | GCN | GAT | HAT |
|---|---|---|---|
| Entire | 0.9007 | 0.9028 | 0.9158 |
| 2 | 0.8827 | 0.8864 | 0.8984 |
| 4 | 0.8753 | 0.8778 | 0.8903 |
| 8 | 0.8654 | 0.8670 | 0.8811 |

*3.2.3 Impact of data distribution.* To investigate how the data distribution impacts the model performance, we study this by varying the proportion (Prop.) of data. The results are shown in Table 3. We find that with the proportion of data held by A and B is even, i.e., from 1:9 to 5:5, the performances of most strategies tend to decrease. This is because the neighbor aggregation is done by data holders individually, and with a larger proportion of data held by a single holder, it is easier for this party to generate better local embeddings.

Table 3: Micro F1 Score of different base GNN models (Dataset: ACM, Strategy: concatenation).

| Ratio | GCN | GAT | HAT |
|---|---|---|---|
| Entire | 0.9007 | 0.9028 | 0.9158 |
| 5:5 | 0.8827 | 0.8864 | 0.8984 |
| 3:7 | 0.8908 | 0.8928 | 0.9059 |
| 1:9 | 0.8998 | 0.9022 | 0.9143 |

## 3.3 Complexity Analysis

Compared with FL, SL is more communication efficient with an increase in the number of participants or model size [24], and a decrease in the scale of the dataset. In contrast, in FL, the communication efficiency will skyrocket, when the number of participants is large, as shown in Figure 2.

## 4 CONCLUSION AND DISCUSSION

In this paper, we have presented a novel approach (SplitGNN), which splits the computation graph of GNN by leaving the private datarelated computations to participants and delegating the rest to the server. SplitGNN outperformed the previous SOTA approach on open-source datasets. We also conducted the communication efficiency (lower communication costs than FL). In addition to HE considered in our work, previous works have applied DP on the local embeddings [16, 17] when the server generates global embeddings, to further protect against information leakage. Compared to encryption-based schemes, DP-based approaches are more computationally efficient [21], however, it may hurt the utility. Therefore, we leave the comparative performance analysis of SplitGNN with the integration of DP for future work.

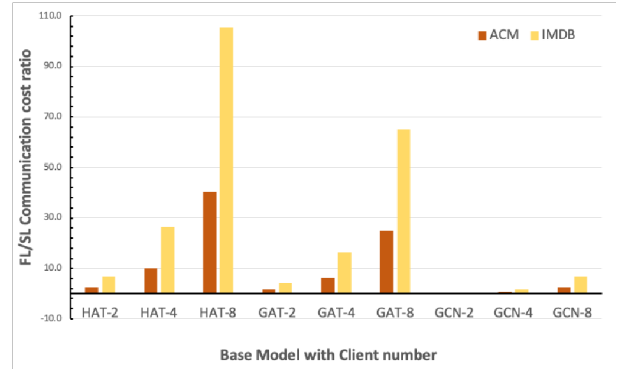

Figure 2: SplitGNN Complexity Analysis.